\documentclass[letterpaper]{article} % DO NOT CHANGE THIS
\usepackage{aaai25}
\usepackage{times}  % DO NOT CHANGE THIS
\usepackage{helvet}  % DO NOT CHANGE THIS
\usepackage{courier}  % DO NOT CHANGE THIS
\usepackage[hyphens]{url}  % DO NOT CHANGE THIS
\usepackage{graphicx} % DO NOT CHANGE THIS
\usepackage{natbib}  % DO NOT CHANGE THIS AND DO NOT ADD ANY OPTIONS TO IT
\usepackage{caption} % DO NOT CHANGE THIS AND DO NOT ADD ANY OPTIONS TO IT
\usepackage{algorithm}
\usepackage{algorithmic}
\usepackage{amsthm,amsmath,amssymb}
\usepackage{natbib}
\usepackage{booktabs}
\usepackage{multirow}
\usepackage{pifont}
\newtheorem{lemma}{Lemma}
\newtheorem{theorem}{Theorem}

\newcommand{\mbbR}{\mathbb{R}}
\newcommand{\mbbP}{\mathbb{P}}

\newcommand{\R}{\mathbb{R}}

\usepackage{newfloat}
\usepackage{listings}
\DeclareCaptionStyle{ruled}{labelfont=normalfont,labelsep=colon,strut=off} % DO NOT CHANGE THIS
\floatstyle{ruled}
\newfloat{listing}{tb}{lst}{}
\floatname{listing}{Listing}
\title{Conformal Inference of Individual Treatment Effects Using Conditional Density Estimates}
\author{
    Baozhen Wang, Xingye Qiao
}
\affiliations{
    Department of Mathematics and Statistics, Binghamton University\\
    $\left\{\text{bwang62, xqiao}\right\}$@binghamton.edu
}

\usepackage{bibentry}
\begin{document}

\maketitle

\begin{abstract}
In an era where diverse and complex data are increasingly accessible, the precise prediction of individual treatment effects (ITE) becomes crucial across fields such as healthcare, economics, and public policy. Current state-of-the-art approaches, while providing valid prediction intervals through Conformal Quantile Regression (CQR) and related techniques, often yield overly conservative prediction intervals. In this work, we introduce a conformal inference approach to ITE using the conditional density of the outcome given the covariates. We leverage the reference distribution technique to efficiently estimate the conditional densities as the score functions under a two-stage conformal ITE framework. We show that our prediction intervals are not only marginally valid but are narrower than existing methods. Experimental results further validate the usefulness of our method.
\end{abstract}


\section{Introduction}\label{introduction}
Understanding the effect of interventions on an individual level is important across many domains, such as healthcare, economics, and public policy. Traditional average treatment effect estimates consider all individuals but fail to account for the heterogeneity in individual responses. As diverse data from various fields become more accessible, machine learning plays an increasingly significant role in revealing insights from the data.

%Precision medicine is an innovative approach that uses personalized and targeted treatment strategies based on an individual's unique characteristics, such as their genetics, environment, and lifestyle. This approach marks a significant difference from the traditional one-size-fits-all model of healthcare, allowing for more precise, effective, and tailored interventions. Machine learning plays an increasingly significant role in precision medicine.

In recent years, many research works have focused on machine learning algorithms that provide point estimates of the Conditional Average Treatment Effect (CATE) \citep{athey2016recursive, wager2018estimation, kunzel2019metalearners, meng2020doubly}, which quantifies the expected difference in treatment outcomes for individuals with specific characteristics. While CATE takes into account the varying covariates of the individuals by averaging effects across similar individuals, it can still sometimes overlook individual-level heterogeneity. Recent works on Individualized Treatment Effect (ITE) have made significant progress in addressing these limitations, offering more accurate predictions tailored to each individual \citep{hill2011bayesian, alaa2017bayesian, shalit2017estimating}. They utilize Gaussian processes or Bayesian approaches to provide interval-valued predictions at the individual level. However, these methods could be model-specific, and not always ensure that the predicted confidence intervals achieve the nominal coverage probability, i.e., the proportion of times that the true treatment effect lies within the constructed intervals across different samples. This issue can limit the reliability and generalizability of the treatment effect estimates, particularly when applied to diverse patient populations or under varying clinical conditions.

Conformal prediction \citep{vovk2005algorithmic, lei2013distribution, lei2018distribution} provides a model-agnostic and distribution-free framework that outputs interval predictions with the desired coverage probability. The main idea of (split) conformal prediction is to compute a conformity score by fitting a predictive model on the training set, and then evaluate the conformity score on the calibration set to quantify the uncertainty of future predictions. Although the conformal framework is model-agnostic and always achieves a coverage guarantee, the average length of resulting prediction intervals can highly rely on the choice of the score functions. \citet{lei2021conformal} propose a two-stage methodology utilizing weighted conformal prediction (WCP) \citep{tibshirani2019conformal} to address the covariate shift problem during counterfactual inference. This method is considered as the state-of-the-art method which provides prediction intervals for ITE problems, with desired coverage guarantee as well as reasonably short intervals. \citet{chen2024conformal} study a similar approach, where they use the joint density to compute weights in WCP. Alternatively, \citet{alaa2023conformal} develop a methodology called conformal meta-learner, which applies conformal prediction directly with imputed pseudo outcomes. Both methods utilize Conformal Quantile Regression (CQR) \citep{romano2019conformalized}, an approach that integrates the concept of conformal prediction with quantile regression. CQR is capable of adapting to any heteroscedasticity in the data, but may not guarantee the shortest prediction intervals. Additionally, the two-stage design inherent in both methods tends to produce conservative results. Experimental studies consistently show that these methods yield conservative prediction intervals with a coverage much greater than the desired level \citep{lei2021conformal, alaa2023conformal}. Motivated by the need for more precise prediction interval for ITE, this work aims to refine these methodologies to achieve shorter prediction intervals while maintaining the coverage guarantee.

Inspired by the conformal predictions under the classification setting \citep{lei2014classification, sadinle2019least}, one can show that directly using the conditional density of the outcome $Y$ given the covariates $X$ as the score function in conformal prediction will lead to the shortest prediction interval (see discussion in Section \ref{subsec:cond}). While a substantial amount of research has focused on estimating the regression function \(E(y|x)\), the task of estimating the full conditional density \(f(y|x)\), particularly in scenarios where \(x\) is high-dimensional, has received considerably less attention. \citet{izbicki2022cd} proposed a framework that utilizes conditional densities under regression setting. However, their focus is on achieving local conditional coverage and they employ a non-parametric smoothing technique to estimate the conditional densities, which is less efficient. 

In this paper, we introduce a novel approach to conformal inference of individual treatment effects (ITE) using conditional densities as the score function. To address the computational difficulties associated with estimating full conditional densities, we employ a reference distribution technique to alleviate the problem. We theoretically show that the proposed method achieves shorter prediction intervals as well as maintaining the desired coverage guarantee. Empirical studies, including both simulations and semi-synthetic benchmarks, strongly indicate that our proposed method surpasses existing state-of-the-art methods in prediction length.

The remainder of this article is outlined as follows. We begin by reviewing the background knowledge of the potential outcome framework and conformal prediction in Section \ref{sec:review}. We introduce our methodologies along with the discussion of theoretical guarantee in Section \ref{sec:method}. In Section \ref{sec:numerical}, we present supporting simulation and semi-synthetic experiments. Section \ref{sec:conclusion} concludes by summarizing our contributions and the practical implications. Proofs and additional discussions can be found in the supplementary material.

\section{Background}\label{sec:review}
In this section, we first introduce the standard potential outcome framework. Then we review the conformal prediction in Section \ref{subsec:conformal} and the weighted conformal prediction under covariate shift in Section \ref{subsec:wconformal}.
\subsection{Potential Outcome Framework and Objective Statement}
We focus on the standard potential outcome framework \citep{neyman1923application,rubin1974estimating} with a binary treatment. Denote by \(A = \left\{0,1\right\}\) the binary treatment indicator, by \(X\in \mathcal{X}\subseteq\mathbb{R}^d\) the covariates, by \(Y\in \mathcal{Y}\subseteq\mathbb{R}\) the observed outcome. For each subject \(i\),  let $Y_i(1)$ and $Y_i(0)$ be the pair of potential outcomes under \(A = 1\) and \(A = 0\) respectively. The fundamental problem of causal inference is that we can only observe one potential outcome out of $Y_i(1)$ and $Y_i(0)$ for each subject \citep{holland1986statistics}. Let
\begin{equation*}
    (X_i, A_i, Y_i(1), Y_i(0))\overset{\text{i.i.d}}{\sim} P(X, A, Y(1), Y(0)).
\end{equation*}
The following assumptions are often considered: (1) \textit{Unconfoundedness (strong ignorability)}: \( (Y(1), Y(0)) \perp A|X\), which allows us to interpret the differences in outcomes as causal effects, rather than being confounded by other factors. Under unconfoundedness, the conditional distributions of a potential outcome are invariant across treatment groups: \(P(Y|X,A=a) = P(Y(a)|X)\). (2) \textit{Stable unit treatment value assumption} (SUTVA): \( Y_i = Y_i(A)\), which ensures individual treatment effects are not influenced by other units. (3) \textit{Positivity}: \( 0< P(A = 1|X = x) < 1\), which ensures that every individual has a nonzero probability of receiving each treatment condition.

Existing methods mostly focused on conditional average treatment effects (CATE) \citep{athey2016recursive, wager2018estimation, kunzel2019metalearners}, defined as \(\tau(x) = E(Y(1) - Y(0)|X= x)\). In this work, our primary focus is on individual treatment effects (ITE), defined as \(Y_i(1) - Y_i(0)\) for subject $i$ without knowing the treatment assignment, and to construct a prediction interval of the ITE. Given the observations \((X_i, Y_i, A_i),  i = 1,\dots,n\), our goal is to construct a predictive interval \(\hat{C}(x)\) that covers the true ITE for a new test individual \(n+1\) with covariate \(X_{n+1}\) with high probability, i.e.
\begin{equation}\label{eq:ITEcov}
    \mbbP\left(Y_{n+1}(1) - Y_{n+1}(0) \in \hat{C}(X_{n+1})\right) \geq 1 - \alpha,
\end{equation}
for a pre-specified level \(\alpha\in(0,1)\). Typically, \(\alpha\) is a small value, such as 0.05.

\subsection{Conformal Prediction}\label{subsec:conformal}
Conformal prediction \citep{vovk2005algorithmic, lei2013distribution, lei2018distribution} provides a means to a prediction set that with a predetermined probability covers the true value for future individuals based on a finite sample. Below we describe the construction of the original conformal prediction set. We first choose a score function $S(\cdot,\cdot)$, whose arguments consist of a point $(x,y)$, and some dataset $D$. By convention, a low value of $S((x,y), D)$ indicates that the point $(x,y)$ ``conforms'' to $D$, whereas a high value indicates that $(x,y)$ is atypical relative to the points in $D$. For convenience of the method development below in this article, we choose $S$ to be a conformity, instead of nonconformity, score; that is, a high value of $S$ indicates that $(x,y)$ ``conforms'' to $D$. 

Given a training data set $(X_i,Y_i), i =1,...,n$, and a fixed $x\in\mbbR^d$, we obtain $\hat{C}(x)\subseteq \mbbR$, the conformal prediction set, by repeating the following procedure for each $y_{\text{trial}}\in\mbbR$: we first calculate the conformity scores $V_i^{(x,y_{\text{trial}})} = S((X_i,Y_i),  \bigcup_{i=1}^n (X_i,Y_i)\cup \left\{ (x,y_{\text{trial}})\right\})$, for  $i = 1,\dots, n,$ and $V_{n+1}^{(x,y_{\text{trial}})} = S((x,y_{\text{trial}}), \bigcup_{i=1}^n (X_i,Y_i)),$ and then we include $y_{\text{trial}}$ in $\hat{C}(x)$ if $V_{n+1}^{(x,y_{\text{trial}})} \geq \text{Quantile} (\alpha; V_{1:n}^{(x,y_{\text{trial}})} \cup \left\{ \infty \right\})$, that is, if no less than $\alpha(n+1)$ many of $V_i^{(x,y_{\text{trial}})}$'s are no greater than $V_{n+1}^{(x,y_{\text{trial}})}$. Importantly, the symmetry in the construction of the conformity scores guarantees a satisfactory coverage rate in finite samples \citep{lei2018distribution}:
 \begin{equation}\label{UncondCov}
 \mbbP (Y \in \hat{C} (X) ) \geq 1 - \alpha,
 \end{equation}
 where $\mbbP$ is taken over the $n+1$ i.i.d. draws of training samples and the test point.

The above original version of conformal prediction provides a finite-sample guarantee of the coverage rate. However, it can be computationally expensive, especially if $S$ has to be computed through an expensive machine-learning method. A more popular alternative is the split-conformal method \citep{papadopoulos2002inductive, vovk2005algorithmic}, where the entire training data is split into two parts. The first part is used to estimate the score function $S$, which is then evaluated on the second part of the data. The splitting process not only alleviates the computational burden of the full conformal prediction but also mitigates the risk of overfitting, as the score function is calibrated on a separate set from where it was trained.

\subsection{Weighted Conformal 
Prediction under Covariate Shift}\label{subsec:wconformal}

Both the original and the split version of conformal prediction assume that the distributions of the target data and the training data are the same. \citet{tibshirani2019conformal} generalized conformal prediction for regression to WCP under covariate shift assumptions. Covariate shift \citep{shimodaira2000improving, sugiyama2007direct} refers to the scenario where the marginal distribution of covariates differs between the training and target data set, denoted as \(P_X\) and \(Q_X\) respectively, while the conditional distributions of the outcome given the covariates remain the same, denoted as \(P_{Y|X}\). \citet{lei2021conformal} studied the counterfactuals inference utilizing WCP under covariate shift, where
\[\text{Training: } (X,Y) \sim P_{X} \times P_{Y|X} ;
\]
\[\text{Target: } (X,Y) \sim Q_{X} \times P_{Y|X}.\] 

Here we briefly go over the algorithm. Assume that the probability measure of the target data covariates is absolutely continuous with respect to that of the training data covariates, we consider using the Radon-Nikodym derivative \(w(x) = dQ_X(x)/dP_X(x)\) to address the shift in covariate distribution. Define \(p_i(x) = w(x_i)/[\sum_{i'=1}^n w(x_{i'})+w(x)]\), for \(i=1,\dots,n\) and \(p_{n+1}(x) = w(x)/[\sum_{i'=1}^n w(x_{i'})+w(x)]\). We can use weighted quantile of the scores computed in the calibration data as the cutoff value, with \(p_i(x)\) as the weight, to obtain the prediction set: 
\begin{equation*}
\begin{aligned}
\hat{C}(x) &= \biggl\{ y \in \R : V_{n+1}^{(x,y)} \\
&\geq \text{Quantile}\biggl(\alpha; \sum_{i=1}^{n} p_i(x) \delta_{V_i^{(x,y)}} + p_{n+1}(x) \delta_{\infty}\biggr)\biggr\},
\end{aligned}
\end{equation*}
where \(\delta_c\) is a Dirac measure placing a point mass at \(c\). Assume we have the true value of \(w(x)\), \citet{tibshirani2019conformal} showed that \(\hat{C}(x)\) satisfies:
\[\mbbP_{(X,Y)\sim Q_X\times P_{Y|X}} (Y\in \hat{C}(X)) \geq 1 -\alpha.\]

Under the potential outcome framework, the covariate distribution of the training data is a mixture of \(P_{X|A=1}\) and \(P_{X|A=0}\). WCP can not directly handle training data of a mixed type due to computational challenges associated with weighted calculations. \citet{lei2021conformal} propose a two-stage framework to overcome this issue. On the first stage, one can use training data from the treatment group \(P_{X|A = 1} \times P_{Y|X}\), to produce interval estimates for those from control group \(P_{X|A = 0} \times P_{Y|X}\) via WCP and vice versa. Then, in the second stage, one can integrate the interval outcomes from both groups using a secondary conformal prediction procedure or a naive Bonferroni correction. In Section \ref{subsec:stage1}, we will explore this two-stage framework, and propose another less conservative alternative using the concept of X-learner \citep{kunzel2019metalearners}.

\section{Methodology}\label{sec:method}
In this section, we begin by demonstrating how using the conditional density as the score function can optimize the length of the conformal prediction interval. In Section \ref{subsec:reference}, we introduce a reference distribution technique for efficient estimation of conditional densities. In Section \ref{subsec:stage1}, we adopt the two-stage framework proposed by \citet{lei2021conformal} to develop algorithms that compute shorter prediction intervals for the Individual Treatment Effect (ITE) of new subjects, ensuring desired coverage guarantees.

\subsection{Conditional Density as the Score Function}\label{subsec:cond}
Let \(\mathbb{P}\) denote the joint distribution of \((X,Y)\) and \(f\) denote the density of \(\mathbb{P}\) with respect to Lebesgue measure. Throughout the article, we denote \(f(y|x)=f(Y=y|X=x)\) as the conditional density of \(Y\) equaling \(y\) given \(X\) equals \(x\).

We define \( C: \mathbb{R}^d \rightarrow \mathcal{M}(\mathbb{R})\), where \(\mathcal{M}(\mathbb{R})\) represents the set of all measurable intervals over \(\mathbb{R}\). The function \( C \) serves as a confidence interval predictor, which provides an interval \( C(x) \) intended to contain the response variable \( y \), based on input \( x \). Consider the following optimization problem that minimizes the expected length of these intervals while ensuring that the probability of \( y \) falling within \( C(x) \) is at least the predefined level \(1-\alpha\):
\begin{equation}\label{eq:obj}
\min_{C}\mathbb{E} \left\{|C(X)| \right\} \ \ \ \text{  subject to  }\ \ \ \mathbb{P}\left\{y \in C(X) \right\} \geq 1 - \alpha
\end{equation}

\begin{theorem}\label{thm1}
Let \(t_\alpha\) denote the \(\alpha\) quantile of \(f(Y|X=x)\). The solution that optimizes \eqref{eq:obj} is given by $C_{\alpha}^* = \left\{(x,y) : f(y|x) \geq t_{\alpha} \right\}$. And the optimal predictor can be written as 
\begin{equation}\label{eq:C}
    C_{\alpha}^*(x) = \left\{y: f(y|x)\geq t_{\alpha}\right\}.
\end{equation}
\end{theorem}

The proof can be found in the supplementary material. A similar problem of \eqref{eq:obj} has been explored under classification contexts by \citet{lei2014classification} and \citet{sadinle2019least}. According to Theorem \ref{thm1}, if we can consistently estimate \(f(y|x)\) and apply it as a score function in conformal prediction, as detailed in Algorithm \ref{alg1}, we can optimize the length of the prediction interval while ensuring a coverage guarantee of at least \(1-\alpha\).

\subsection{Estimate Conditional Density Using Reference Distribution Technique}\label{subsec:reference}

Using \(f(y|x)\) as the score function in conformal prediction can be optimal; however, estimating the full conditional density \(f(y|x)\) presents significant challenges. Traditional methods like the non-parametric kernel density estimator \citep{de2003conditional} must address each dimension of \(x \in \mathbb{R}^d\) and often struggle due to the curse of dimensionality. Contemporary approaches, such as tree-based \citep{holmes2012fast} and neural network methods \citep{rothfuss2019conditional}, involve complex computations and may prove less efficient for large-scale applications. To effectively address this issue, we adapt a reference distribution technique originally intended for unconditional density estimation \citep{hastie2009elements}. The details of this adaptation are described as follows.

Suppose we have $n$ i.i.d. random samples drawn from the joint density \(f(x, y) = f(y|x) h(x)\), denoted as \((x_1, y_1), (x_2, y_2), \ldots, (x_n, y_n)\). We use a reference probability density function \(f_0(y)\), from which a sample of size $n$ independent of \(h(x)\) is drawn using Monte Carlo methods, denoted as \( \tilde{y}_1, \tilde{y}_2, \ldots, \tilde{y}_n\). We then combine a duplicate of \(x_1, x_2, \ldots, x_n\) with these \(\tilde{y}\) to form a joint reference distribution \(f_0(x, y) = f_0(y)h(x)\). By assigning \(Z = 1\) to each data point from \(f(x, y)\) and \(Z = 0\) to those from \(f_0(x, y)\), we estimate \(\mu(x, y) := E(Z|x, y)\) by supervised learning using the aggregated dataset:
\begin{equation}\label{eq:mu}
     \mu(x, y) = \frac{f(x, y)}{f(x, y) + f_0(x, y)} = \frac{f/f_0}{1 + f/f_0},
\end{equation}

The resulting estimate, \(\hat\mu(x, y)\), can be inverted to provide an estimate for the joint density
\[
\hat{f}(x, y) = f_0(x, y) \cdot \frac{\hat{\mu}(x, y)}{1 - \hat{\mu}(x, y)}.
\]
Dividing \(h(x)\) on both sides, we obtain an estimator of conditional density:
\begin{equation}\label{eq:cd}
    \hat{f}(y|x) = f_0(y) \cdot \frac{\hat{\mu}(x, y)}{1 - \hat{\mu}(x, y)}.
\end{equation}

Techniques such as logistic regression and random forests, which efficiently estimate log-odds \(\log(f/f_0)\), are natural choices for this procedure. Generally, many reference density can be used for $f_0(y)$, provided that the support of the reference density covers that of the original one. However, in practice, the accuracy of \(\hat{f}(y|x)\) can be influenced by the choice of $f_0(y)$. We recommend using a Gaussian distribution with the same mean and a slightly larger variance than the original \(y\)'s to alleviate the extreme case of none overlapping.

\subsection{Weighted Conformal Inference Using Conditional Density Estimates}\label{subsec:stage1}

By leveraging the reference distribution technique to efficiently estimate the conditional density, we can apply the weighted conformal prediction \citep{tibshirani2019conformal} to derive an estimate of \eqref{eq:cd}. As introduced in Section \ref{subsec:wconformal}, \citet{lei2021conformal} propose a two-stage framework to overcome the challenges of WCP when training data exhibit a mixed distribution of covariates under the potential outcome framework. A two-stage framework seems a common and necessary choice to compensate for never observing the potential outcome. \citet{alaa2023conformal} also utilize a two-stage framework, where they first impute a pseudo outcome and then apply conformalized quantile regression (CQR) on these pseudo outcomes along with the covariates. A central motivation of this paper is to reduce the length of the prediction interval as much as possible, due to the inherent conservativeness of prediction intervals under the potential outcome framework.

Algorithm \ref{alg1} below outlines the procedure of the first stage, where we implement WCP using the conditional density estimate $\hat{f}(Y|X)$ as the score function. Within the first stage of our framework, Algorithm 1 is implemented twice: once using training data from the treatment group to obtain interval estimates for the control group, and conversely, using control group data to estimate intervals for the treatment group. In the former scenario, the weight function \(w(x)\) is calculated as:
\begin{equation}\label{eq:w1}
     \frac{dP_{X|A=0}(x)}{dP_{X|A=1}(x)} \propto\frac{P(A=0|X=x)}{P(A=1|X=x)} = \frac{1-\pi(x)}{\pi(x)}
\end{equation}
where \(\pi(x) := P(A=1|X=x)\) is the propensity score \citep{rosenbaum1983central}. This score captures the treatment assignment mechanism under given covariate conditions.

Similarly, when using data from the control group to estimate intervals for the treatment group, the weight function is:
\begin{equation}\label{eq:w0}
    w(x) = \frac{dP_{X|A=1}(x)}{dP_{X|A=0}(x)} \propto\frac{\pi(x)}{1-\pi(x)}
\end{equation}
Upon implementing Algorithm \ref{alg1} in both scenarios, we first partition the training data into two splits, indexed by \(\mathcal{I}_{1}\) and \(\mathcal{I}_{2}\). The first split is used to estimate the weight function \(\hat{w}(x)\) and the conditional density estimate \(\hat{f}(Y|X)\) using the reference distribution technique. We then evaluate both \(\hat{w}(x)\) and \(\hat{f}(Y|X)\) on the second split, then compute the threshold \(\hat{t}_\alpha\) as the \(\alpha\) quantile of the weighted conformity scores.

\begin{algorithm}
\caption{Weighted Split-Conformal using Conditional Density Estimate (CD)}
\label{alg1}
\begin{algorithmic}
\STATE \textbf{Input:} Level $\alpha$, data $(X_i, Y_i)$ from $P_X\times P_{Y|X}$ where $i\in\mathcal{I}$, and a test point $X_{n+1}$ from $Q_X$
\STATE \textbf{Output:} A prediction set $\hat{C}(x)$
\end{algorithmic}
\begin{algorithmic}[1]
\STATE Split $\mathcal{I}$ into two equal sized subsets $\mathcal{I}_{1}$ and $\mathcal{I}_{2}$.
\STATE Estimate the weight function $\hat{w}(x)$ using $\mathcal{I}_{1}$.
\STATE For each $i \in \mathcal{I}_{1}$, generate $\tilde{Y}_i$ from a normal distribution with the same mean and slightly larger variance of data in $\mathcal{I}_{1}$.  Assign $Z = 1$ to $(X_i, Y_i)$ and $Z = 0$ to $(X_i, \tilde{Y}_i)$, then fit a classification algorithm $\hat\mu$ to obtain $\hat{f}(Y|X)$ according to Section \ref{subsec:reference}.
\STATE For each $i \in \mathcal{I}_{2}$, compute the score $V_i = \hat{f}(Y_i|X_i)$ and the weight $\hat{w}(X_i)$
\STATE Compute the normalized weights $\hat{p}_i(x) = \hat{w}(X_i)/\bigl[\sum_{i\in \mathcal{I}_{2}} \hat{w}(X_i) + \hat{w}(X_{n+1})\bigr]$ and $\hat{p}_{n+1} (x) = \hat{w}(X_{n+1})/\bigl[\sum_{i\in \mathcal{I}_{2}} \hat{w}(X_i) + \hat{w}(X_{n+1})\bigr]$
\STATE Compute $\hat{t}_{\alpha}$ as the $\alpha$-th quantile of the distribution $\sum_{i\in\mathcal{I}_{2}} \hat{p}_i(x) \delta_{V_i} + \hat{p}_\infty (x) \delta_\infty$
\RETURN $\hat{C}(X_{n+1}) = \bigl\{y: \hat{f}(y|X_{n+1}) \geq \hat{t}_{\alpha}\bigr\}$
\end{algorithmic}
\end{algorithm}

\begin{theorem}\label{thm2}
Under Algorithm 1, if the non-conformity scores \(V_i\) have no ties almost surely, \(Q_X\) is absolutely continuous with respect to \(P_X\), and \(\mathbb{E}[\hat{w}(X)] < \infty\), then, given the true weights, i.e., \(\hat{w}(\cdot) = w(\cdot)\):
    \[
     P_{(X,Y) \sim Q_X \times P_{Y|X}}(Y_{n+1} \in \hat{C}(X_{n+1})) \geq 1 - \alpha.
    \]
In general, if \(\hat{w}(\cdot) \neq w(\cdot)\), define \(\Delta w = \frac{1}{2} \mathbb{E}_{X \sim P_X}|\hat{w}(X) - w(X)|\). In this case, coverage is lower bounded by \(1 - \alpha - \Delta w\).
\end{theorem}

Theorem \ref{thm2} establishes that for any choice of target covariate distribution \(Q_X\), it is possible to obtain a prediction interval for the outcome with a desired coverage guarantee. The first part of Theorem \ref{thm2} is a split version of Theorem 2 from \citep{tibshirani2019conformal} and the second part adapts from Theorem 3 in \cite{lei2021conformal}. We provide proofs in the supplementary material for completeness.

In stage one, we implement Algorithm \ref{alg1} twice: once using training data from the treatment group in \(\mathcal{I}_1\) with weights defined in \eqref{eq:w1} to compute \(\hat{C}_i(X_i)\) for \(i\in\mathcal{I}_2\) where \(A_i = 0\), and once using data from the control group in \(\mathcal{I}_1\) with weights in \eqref{eq:w0} to compute \(\hat{C}_i(X_i)\) for \(i\in\mathcal{I}_2\) where \(A_i = 1\). We obtain prediction sets \(\hat{C}_i(X_i)\) for each point in \(\mathcal{I}_2\), which satisfy
\begin{equation}\label{eq:covYA}
    \mbbP (Y_i(1 - j)\in \hat{C}_i(X_i)|A_i = j) \geq 1 -\alpha.
\end{equation}
Recall for each individual \(i\) with \(A_i = j\), we can observe its factual outcome \(Y_i = Y_i(j)\). That means we can obtain a confidence interval of $\text{ITE}_i = Y_i(1) - Y_i(0)$ for each \(i\in\mathcal{I}_2\) by subtraction. Specifically, for those in the treatment group, i.e., \(A_i = 1\), 
\begin{equation}\label{eq:C1}
\hat{C}_i = \bigl[Y_i(1) - \max(\hat{C}(X_i)), Y_i(1) - \min(\hat{C}(X_i))\bigr], 
\end{equation}
and for those in the control group, i.e., \(A_i = 0\), 
\begin{equation}\label{eq:C0}
\hat{C}_i = \bigl[\min(\hat{C}(X_i)) - Y_i(0), \max(\hat{C}(X_i)) - Y_i(0) \bigr]. 
\end{equation}
Eq. \eqref{eq:C1} and \eqref{eq:C0} are useful, but they depend on knowing the value of $A_i$, which is not available for future data. 

In stage two, we apply a secondary procedure to the new training data pairs \((X_i, \hat{C}_i)\) to eliminate the dependency on the treatment assignment \(A\). Algorithm \ref{alg2} below sketches the \textit{Exact} method, where we apply another split Conformal on \((X_i, \hat{C}_i)\). Here we denote $\hat{C}_i = \bigl(\hat{C}^L_i, \hat{C}^U_i\bigr)$ for simplicity.

%, and functions $\hat{\tau}_L(x)$ and $\hat{\tau}_U(x)$ to fit conditional mean of $\hat{C}^L_i$ and $\hat{C}^U_i$

\begin{algorithm}
\caption{CD-Exact}
\begin{algorithmic}\label{alg2}
\STATE \textbf{Input:} Level $\gamma$, data $(X_i, \hat{C}_i),i\in \mathcal{I}_2,$ where $\hat{C}_i$ are obtained from Algorithm \ref{alg1} using control and treatment group respectively, as shown in \eqref{eq:C1} and \eqref{eq:C0}, and a test point $X_{n+1}$
\STATE \textbf{Output:} A prediction interval $\hat{C}_{\text{ITE}}(X_{n+1})$
\end{algorithmic}
\begin{algorithmic}[1]
\STATE Split $\mathcal{I}$ into two equal sized subsets $\mathcal{I}_{tr}$ and $\mathcal{I}_{ca}$
\STATE On $\mathcal{I}_{tr}$, fit conditional mean $\hat{\tau}_L$ and $\hat{\tau}_U$ of $\hat{C}^L_i$ and $\hat{C}^U_i$ given $X$
\STATE For each $i\in\mathcal{I}_{ca}$, compute score 
$V_i = \max \bigl\{\hat{\tau}_L(X_i) - \hat{C}^L_i, \hat{C}^U_i - \hat{\tau}_U(X_i) \bigr\}$.
\STATE Compute $\eta$ as the $1-\gamma$ quantile over scores $V_i$.
\RETURN $\hat{C}_{\text{ITE}}(X_{n+1}) = \bigl[\hat{\tau}_L(X_{n+1}) - \eta, \hat{\tau}_U(X_{n+1}) + \eta\bigr]$  
\end{algorithmic}
\end{algorithm}

\begin{lemma}[\citet{lei2021conformal}]\label{lemma1}
  Assume $(X_i,\hat{C}_i)$ are i.i.d. from $(X, C)$. Then, for a test point \(X_{n+1}\) under Algorithm \ref{alg1} and \ref{alg2}, both with miscoverage level \(\alpha/2\), 
  \[\mathbb{P}(Y_{n+1}(1)-Y_{n+1}(0)\in \hat{C}_{\text{ITE}}(X_{n+1}) \geq 1 - \alpha.\]
\end{lemma}

Lemma \ref{lemma1} can be directly proved using Theorem \ref{thm2} above and Theorem 2 in \citep{lei2021conformal}. Lemma \ref{lemma1} shows that by using an exact method, we can construct a prediction interval that covers the true ITE with a desired coverage level as in \eqref{eq:ITEcov}. 

Another method with theoretical guarantees is the \textit{Naive} method, which employs a straightforward approach using a naive Bonferroni correction, designed as follows:
\begin{equation*}
\begin{aligned}
\hat{C}_{\text{ITE}}(x) = \bigl[&\min(\hat{C}^1(x)) - \max(\hat{C}^0(x)), \\
&\max(\hat{C}^1(x)) - \min(\hat{C}^0(x)) \bigr]
\end{aligned}
\end{equation*}
Here \(\hat{C}^1\) and \(\hat{C}^0\) denote the prediction intervals computed for the treatment group and the control group, respectively. By setting the miscoverage level in Algorithm \ref{alg1} to be \(1-\alpha/2\), the Naive method achieves the desired coverage probability of \(1 - \alpha\) for the ITE, as specified in \eqref{eq:ITEcov}.

In practice, while the Exact and Naive methods tend to be overly conservative, utilizing the conditional density estimate as the score function in Algorithm \ref{alg1} helps to mitigate this issue, as demonstrated in our empirical results in Section \ref{subsec:simu}. A practical and favorable alternative is the \textit{Inexact} method, which yields much shorter prediction intervals, albeit without theoretical guarantees. To implement the Inexact method, we fit plug-in estimates for the 40\% conditional quantiles of \(\hat{C}^L_i\) and 60\% conditional quantiles of \(\hat{C}^U_i\), respectively. For a new test point, these quantiles are then used to straightforwardly compute the prediction interval.

Inspired by X-learner \citep{kunzel2019metalearners}, we propose a fourth, less conservative alternative method named \textit{CD-X}. We fit four plug-in estimates for the conditional means of \(\hat{C}^L_i\) and \(\hat{C}^U_i\) for both \(A_i = 0\) and \(A_i = 1\), denoted as \(\tilde{C}^L_0(x), \tilde{C}^U_0(x), \tilde{C}^L_1(x), \tilde{C}^U_1(x)\). In Algorithm \ref{alg1}, we also estimate the propensity scores \(\hat{\pi}(x)\) using \(\mathcal{I}_1\). Then, for a new test individual, the prediction interval can be computed using the formula
\begin{equation*}
\begin{aligned}
\hat{C}_{\text{ITE}}(x) = \bigl[ &\hat{\pi}(x)\tilde{C}^L_1(x)+(1-\hat{\pi}(x))\tilde{C}^L_0(x), \\
&\hat{\pi}(x)\tilde{C}^U_1(x)+(1-\hat{\pi}(x))\tilde{C}^U_0(x)\bigr]
\end{aligned}
\end{equation*}
Although CD-X does not offer theoretical guarantees, it performs well in most cases within our numerical experiments, achieving the desired coverage meanwhile producing the shortest prediction intervals. In the next section, we will examine the empirical performance of methods ensemble with CD (Algorithm \ref{alg1}) and WCP.

\begin{figure*}[t]
    \centering
    \includegraphics[width=1\linewidth]{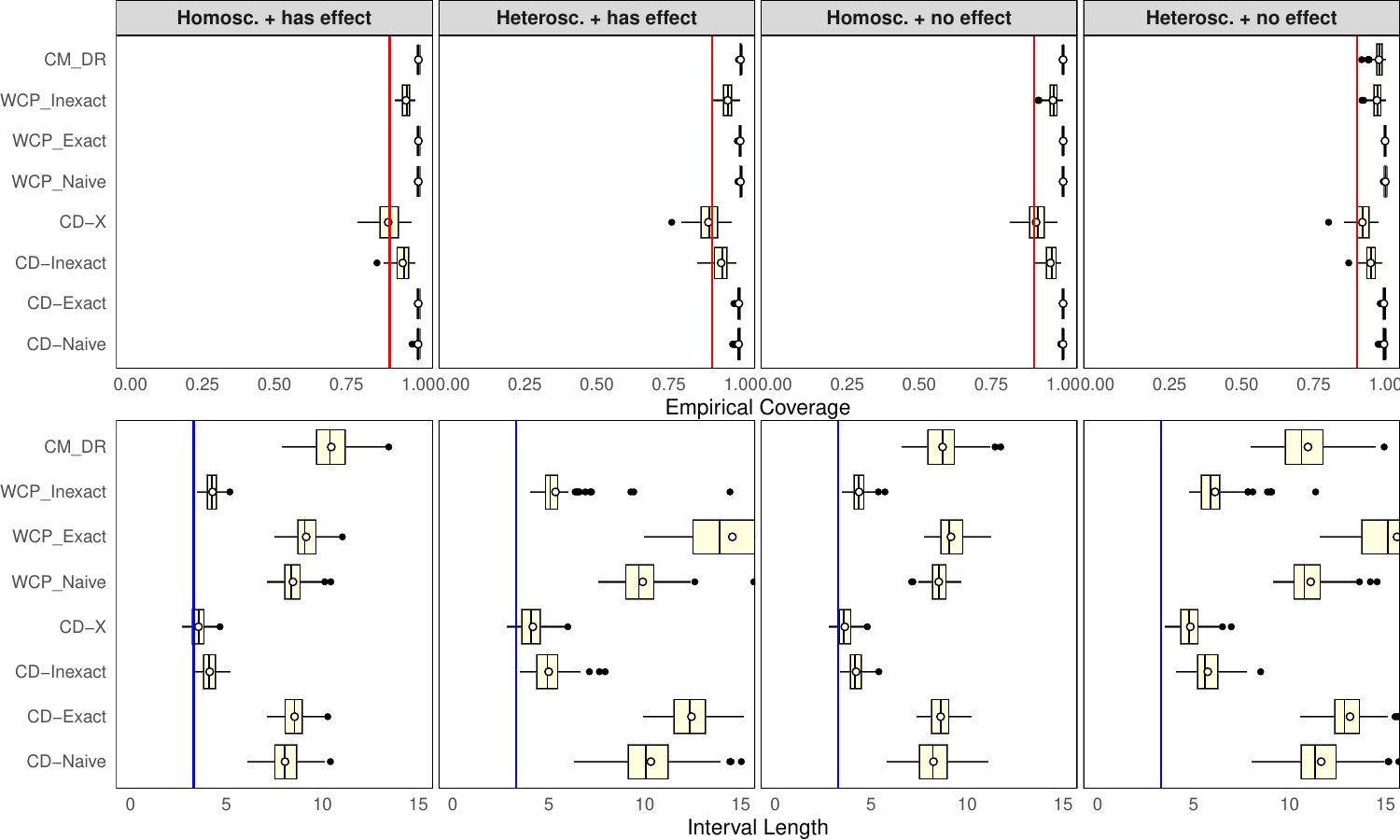}
    \caption{Performance of all baselines in four simulation scenarios described in Section \ref{subsec:simu}. The red vertical lines correspond to target coverage, and the blue vertical lines correspond to the optimal interval width. Here CD stands for Algorithm \ref{alg1}, WCP stands for weighted conformal prediction, and CM stands for conformal meta-learners.}
    \label{fig:1}
\end{figure*}

\section{Experimental Studies}\label{sec:numerical}
\subsection{Experimental Setup}
The nature of potential outcomes limits our observations to factual outcomes, excluding counterfactuals. This characteristic necessitates the validation of ITE estimation primarily through simulations and semi-synthetic data. In this section, we will explore the performance of our methods across one simulation featuring four different settings, and three semi-synthetic benchmarks.

We consider the following baselines known to produce valid prediction intervals for ITEs:
\begin{itemize}
    \item \textbf{WCP-Inexact}, \textbf{WCP-Exact}, and \textbf{WCP-Naive}: We consider state-of-the-art (SOTA) methods based on WCP, each integrating the Inexact, Exact, and Naive approaches.
    \item \textbf{CM-DR}: We also evaluate the Conformal Meta-learners \citep{alaa2023conformal} with the doubly-robust learner. This method is one of the top performers among three methods proposed in \citep{alaa2023conformal} and also offers theoretical guarantees. Throughout our experimental studies, we estimate the propensity score in CM-DR rather than assuming it is known, as done in the original study by \citep{alaa2023conformal}, to ensure a fair comparison.
\end{itemize}

\begin{table*}[t]
\centering
\begin{small}
\begin{tabular}{l|cclcccc|c}
\hline
\multicolumn{1}{c|}{\multirow{2}{*}{Method}} & \multicolumn{3}{c}{NSLM}                     & \multicolumn{2}{c}{ACIC} & \multicolumn{2}{c|}{IHDP} & \multicolumn{1}{c}{\multirow{2}{*}{With guarantee}} \\ \cline{2-8}
\multicolumn{1}{c|}{}                   & Coverage     & \multicolumn{2}{c}{Avg. len.}  & Coverage   & Avg. len.   & Coverage    & Avg. len.   & \multicolumn{1}{c}{}                                \\ \hline
CM-DR                                   &   99.9 (0.00)& \multicolumn{2}{c}{6.45 (0.14)}& 99.9 (0.02)&  52.9 (1.80)&  99.9 (0.03)&  82.1 (5.31)&      \ding{52}                                               \\
WCP-Inexact                             &   93.6 (0.29)& \multicolumn{2}{c}{2.20 (0.03)}& 96.1 (0.48)&  16.6 (0.40)&  83.2 (0.87)&  8.86 (0.72)&      \ding{56}                                               \\
WCP-Exact                               &   99.9 (0.01)& \multicolumn{2}{c}{4.48 (0.04)}& 99.8 (0.04)&  40.7 (2.11)&  99.5 (0.10)&  34.9 (4.60)&      \ding{52}                                               \\
WCP-naive                               &   99.9 (0.01)& \multicolumn{2}{c}{4.23 (0.03)}& 99.8 (0.04)&  30.7 (1.38)&  99.3 (0.13)&  21.4 (2.09)&      \ding{52}                                               \\ \hline
CD-X                                    &   86.3 (0.37)& \multicolumn{2}{c}{1.78 (0.02)}& 93.9 (0.42)&  \textbf{12.2} (0.30)&  79.9 (0.83)&  6.40 (0.51)&      \ding{56}                                               \\
CD-Inexact                              &   92.0 (0.29)& \multicolumn{2}{c}{\textbf{2.09} (0.02)}& 94.7 (0.49)&  14.4 (0.37)&  80.3 (0.93)&  8.31 (0.69)&      \ding{56}                                               \\
CD-Exact                                &   99.9 (0.01)& \multicolumn{2}{c}{4.13 (0.03)}& 99.5 (0.07)&  30.4 (0.63)&  99.4 (0.10)&  25.6 (2.95)&      \ding{52}                                               \\
CD-Naive                                &   99.8 (0.02)& \multicolumn{2}{c}{4.02 (0.04)}& 99.2 (0.13)&  29.3 (0.67)&  96.6 (0.39)&  \textbf{20.2} (2.04)&      \ding{52}                                               \\ \hline
\end{tabular}
\end{small}
\caption{Performance of all methods on semi-synthetic datasets. Empirical coverage (in percentages) and average interval lengths are shown, with standard errors in parentheses. Bold numbers highlight the best performance. ``With guarantee" indicates methods that provide theoretical coverage guarantees.}
\end{table*}

\subsection{Simulation Studies}\label{subsec:simu}
In the simulation studies, we combine the data-generation processes described in \citet{lei2021conformal} and \citet{alaa2023conformal}, which were originally proposed by \citet{wager2018estimation}. The data are generated as follows: Covariates \(X\) are sampled from \(\text{Unif}([0,1]^d)\). The propensity score is generated based on \(\pi(x) = \frac{1}{4} \left[1+\beta_{2,4}(x_1)\right]\), and the treatment assignment \(A|X\) is generated from \(\text{Bern}(\pi(x))\). The potential outcomes, \(Y(j)\) for \(j = 0, 1\), are modeled based on the function \(g(x) = 2/\left\{1+\exp\left[-12(x-0.5)\right]\right\}\), where \(\mathbb{E}[Y(1)|X] = g(x_1)g(x_2)\) and \(\mathbb{E}[Y(0)|X] = \gamma g(x_1)g(x_2)\). Here \(\gamma\) controls the treatment effect. These outcomes are then generated from the model \(\mathbb{E}[Y(j)|X] + \sigma(X)\epsilon\), where \(\epsilon \sim N(0,1)\). We consider four scenarios, derived from a 2 x 2 factorial design: homoscedastic (\(\sigma(x) = 1\)) and heteroscedastic (\(\sigma(x) = -\log (x_1)\)) errors, and treatment has no effect (\(\gamma = 1\)) and the effects are heterogeneous (\(\gamma = 0\)).

\subsection{Semi-synthetic Datasets}
We also explore the performance of our approaches on three semi-synthetic datasets, which feature real covariates combined with simulated outcomes. These datasets include the National Study of Learning Mindsets (NSLM) \citep{yeager2019national}, the 2016 Atlantic Causal Inference Conference Competition (ACIC) \citep{dorie2019automated}, and the Infant Health and Development Program (IHDP) datasets \citep{hill2011bayesian}. Detailed descriptions of all datasets are available in the supplementary material.

\subsection{Results and Discussion}
As our focus is on the prediction interval for ITEs, we use two commonly used metrics across all experimental studies: empirical coverage for true ITEs and average prediction interval length. The empirical coverage is defined as the empirical probability that the true ITE falls within the predicted interval. The average prediction interval length is measured as the mean of the widths of these intervals across all test instances.

Figure 1 illustrates the outcomes of the simulation studies, averaging results over 100 replications for each scenario as described in Section \ref{subsec:simu}. CD-X generally excels, achieving the shortest and near-optimal interval lengths while maintaining the desired coverage levels, except in the scenario with heteroscedastic errors and treatment effect where it undercovers by 0.01. When comparing methods incorporating CD with those using WCP, CD methods consistently provide shorter intervals. Specifically, CD-Inexact is on average 0.26 shorter than WCP-Inexact, CD-Exact is 1.43 shorter than WCP-Exact. CD-Naive and WCP-Naive achieve nearly the same interval lengths on average.

The semi-synthetic experiments further demonstrate the superiority of the proposed CD methods. Across all three benchmarks, CD methods consistently outperform others. Although CD-X and CD-Inexact lack coverage guarantees, they excel in the ACIC and NSLM datasets, respectively, with the shortest average length and the coverage guarantee. For each pair of matched CD and WCP methods, CD consistently provides shorter intervals in \textit{all} cases. Regarding the IHDP results, \citet{alaa2017bayesian} reported that CM-DR performed well with an average interval length of 16.7, assuming a known propensity score. We emphasize that in our studies, we estimate the propensity score for CM-DR to ensure a fair comparison. This is particularly critical given the imbalanced setting of the IHDP dataset, which includes only 747 samples (139 treated and 608 control), making accurate propensity score estimation challenging. In this context, CD-Naive outperforms all other methods.

Although our primary goal is to mitigate the excessive conservatism of prediction intervals, the inherent challenges posed by unobserved counterfactuals naturally lead to conservative outcomes. In our experimental studies, the CD-Exact method consistently demonstrated coverage rates exceeding 99\%, despite a targeted desired coverage of 90\%. Our two-stage framework for predictive inference of ITE addresses the reduction of prediction interval length for counterfactuals in Stage 1 by efficiently using conditional density as the score function. However, further reducing conservatism in Stage 2 remains an area for future research, aiming to find an optimal balance between the less conservative Inexact methods and the overly conservative Exact methods.

\section{Conclusions}\label{sec:conclusion}

In this paper, we developed a two-stage framework to provide interval estimates for Individual Treatment Effects (ITEs), a task inherently challenging due to the nature of unobservable potential outcomes. We successfully leveraged the reference distribution technique to efficiently estimate the optimal conformal scores—the conditional densities. Both theoretical and experimental results demonstrate that our framework outperforms existing state-of-the-art Weighted Conformal Prediction (WCP) methods. The practical implications of our work can be substantial, particularly given the difficulty inherent in estimating ITEs. Our method's success in reducing the average length of prediction intervals enhances their usability in real-world scenarios, particularly in decision-making processes requiring precise estimates, such as in clinical and policy-making fields.

\bibliography{aaai25}

\end{document}